\documentclass[journal]{IEEEtran}
\usepackage{amsmath,amsfonts}
\usepackage{algorithmic}
\usepackage{algorithm}
\usepackage{array}
\usepackage[caption=false,font=normalsize,labelfont=sf,textfont=sf]{subfig}
\usepackage{textcomp}
\usepackage{stfloats}
\usepackage{url}
\usepackage{verbatim}
\usepackage{graphicx}
\usepackage{cite}
\usepackage{multirow}
\usepackage{booktabs}
\usepackage{cuted}
\usepackage{capt-of}
\usepackage{bm}
\usepackage{xcolor}
\usepackage{array}
\usepackage{makecell}
\usepackage[switch,pagewise]{lineno}
\usepackage{scalerel}
\usepackage{tikz}
\usetikzlibrary{svg.path}
\definecolor{orcidlogocol}{HTML}{A6CE39}
\tikzset{
  orcidlogo/.pic={
    \fill[orcidlogocol] svg{M256,128c0,70.7-57.3,128-128,128C57.3,256,0,198.7,0,128C0,57.3,57.3,0,128,0C198.7,0,256,57.3,256,128z};
    \fill[white] svg{M86.3,186.2H70.9V79.1h15.4v48.4V186.2z}
                 svg{M108.9,79.1h41.6c39.6,0,57,28.3,57,53.6c0,27.5-21.5,53.6-56.8,53.6h-41.8V79.1z M124.3,172.4h24.5c34.9,0,42.9-26.5,42.9-39.7c0-21.5-13.7-39.7-43.7-39.7h-23.7V172.4z}
                 svg{M88.7,56.8c0,5.5-4.5,10.1-10.1,10.1c-5.6,0-10.1-4.6-10.1-10.1c0-5.6,4.5-10.1,10.1-10.1C84.2,46.7,88.7,51.3,88.7,56.8z};
  }
}
\newcommand\orcidicon[1]{\href{https://orcid.org/#1}{\mbox{\scalerel*{
\begin{tikzpicture}[yscale=-1,transform shape]
\pic{orcidlogo};
\end{tikzpicture}
}{|}}}}
\usepackage{hyperref}
\hypersetup{
    colorlinks=true,
    linkcolor=blue,
    filecolor=black,      
    urlcolor=black,
    citecolor=blue
}
\usepackage{tikz-network}

\newcommand{\etal}{\textit{et al.}}

\begin{document}

\title{Single-Frame Point-Pixel Registration via Supervised Cross-Modal Feature Matching}

\author{
Yu Han$^{\orcidicon{0009-0006-7656-5835}\,}$,
Zhiwei Huang$^{\orcidicon{0009-0008-7084-052X}\,}$,
Yanting Zhang$^{\orcidicon{0000-0001-6317-1956}\,}$,
Fangjun Ding$^{\orcidicon{0009-0006-7045-8131}\,}$,\\
Shen Cai$^{\orcidicon{0000-0001-5217-3155}\,}$,
Xiaoyu Tang$^{\orcidicon{0000-0002-6038-9623}\,}$,
Yanchao Dong$^{\orcidicon{0000-0001-6864-8354}\,}$,
and Rui Fan$^{\orcidicon{0000-0003-2593-6596}\,}$,~\IEEEmembership{Senior Member,~IEEE}

%
\thanks{\emph{Yu Han and Zhiwei Huang are co-first authors.} \emph{Corresponding author: Yanting Zhang and Rui Fan.}}

\thanks{Yu Han, Yanting Zhang, Fangjun Ding, and Shen Cai are with the School of Information and Intelligent Science, Donghua University, Shanghai 201620, China (e-mails: 2232816@mail.dhu.edu.cn, ytzhang@dhu.edu.cn, 220800613@mail.dhu.edu.cn, cs@dhu.edu.cn).}

\thanks{Zhiwei Huang and Yanchao Dong are with the Department of Control Science \& Engineering, the College of Electronics \& Information Engineering, Tongji University, Shanghai 201804, China (e-mails: \{2431985, dongyanchao\}@tongji.edu.cn).}

\thanks{Xiaoyu Tang is with the School of Electronics and Information Engineering, and Xingzhi College, South China Normal University, Shanwei, 516600, China (e-mail: tangxy@scnu.edu.cn).}

\thanks{Rui Fan is with the Department of Control Science \& Engineering, the College of Electronics \& Information Engineering, Shanghai Research Institute for Intelligent Autonomous Systems, the State Key Laboratory of Intelligent Autonomous Systems, and Frontiers Science Center for Intelligent Autonomous Systems, Tongji University, Shanghai 201804, China, as well as with the National Key Laboratory of Human-Machine Hybrid Augmented Intelligence, Institute of Artificial Intelligence and Robotics, Xi'an Jiaotong University, Xi'an 710049, Shaanxi, China (e-mail: rui.fan@ieee.org).}

}

\maketitle

\begin{abstract}
\label{sec:abstract}
Point-pixel registration between LiDAR point clouds and camera images is a fundamental yet challenging task in autonomous driving and robotic perception. A key difficulty lies in the modality gap between unstructured point clouds and structured images, especially under sparse single-frame LiDAR settings. Existing methods typically extract features separately from point clouds and images, then rely on hand-crafted or learned matching strategies. This separate encoding fails to bridge the modality gap effectively, and more critically, these methods struggle with the sparsity and noise of single-frame LiDAR, often requiring point cloud accumulation or additional priors to improve reliability. Inspired by recent progress in detector-free matching paradigms, we revisit the projection-based approach and introduce the detector-free framework for direct point-pixel matching between LiDAR and camera views. To further enhance matching reliability, we introduce a repeatability scoring mechanism that acts as a soft visibility prior. This guides the network to suppress unreliable matches in regions with low intensity variation, improving robustness under sparse input. Extensive experiments on KITTI, nuScenes, and MIAS-LCEC-TF70 benchmarks demonstrate that our method achieves state-of-the-art performance, outperforming prior approaches on nuScenes (even those relying on accumulated point clouds), despite using only single-frame LiDAR. 

\end{abstract}

\def\abstractname{Note to Practitioners}
\begin{abstract}
In many practical robotic and autonomous driving systems, only sparse LiDAR sensors (e.g., 16- or 32-line) are available due to cost and deployment constraints. This sparsity makes online LiDAR-camera calibration particularly challenging, since traditional methods often rely on dense multi-frame point clouds or additional calibration targets. Our proposed detector-free framework directly establishes point-pixel correspondences from a single sparse LiDAR scan, ensuring robust calibration even under noisy and low-density conditions. This enables practical online re-calibration after sensor maintenance or accidental impacts without requiring specialized setups. Furthermore, the approach allows camera localization against previously recorded sparse point cloud maps, making it valuable for infrastructure perception, historical-map-based navigation, and deployment on low-cost robotic platforms.

\end{abstract}


\begin{figure}
    \centering
    \includegraphics[width=0.95\linewidth]{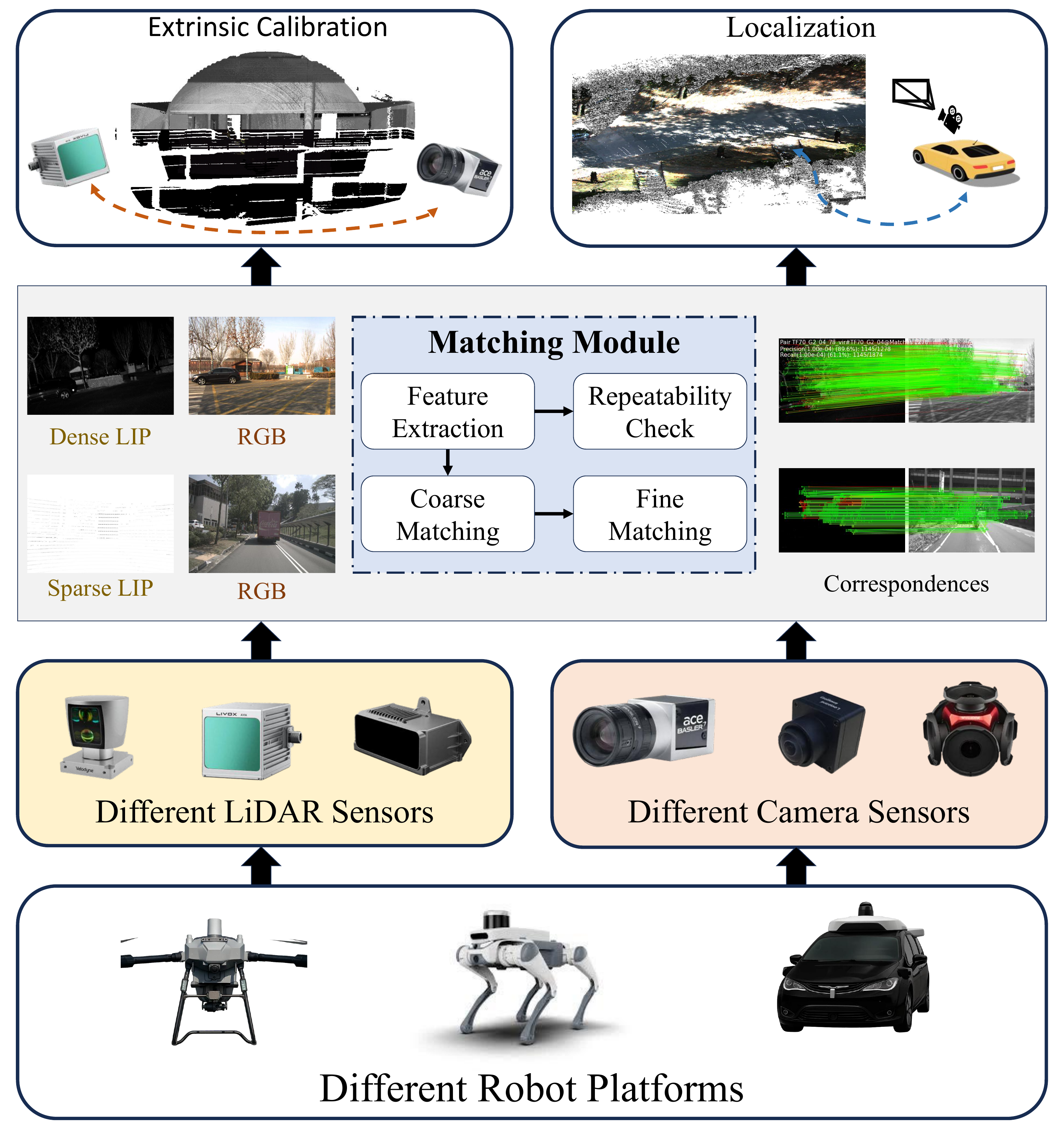}
    \caption{Our method matches dense or sparse projections from various LiDAR sensors with images and recovers the pose via PnP, supporting camera localization in historical point clouds and LiDAR-camera extrinsic calibration.}
    \label{fig:cover}
\end{figure}

\section{Introduction}
\label{sec:intro}
\IEEEPARstart{T}{he} perception is a critical challenge for autonomous robots \cite{tang2024ticoss, chen2024joint}. As the front-end module of a robotic system, perception provides essential information for high-level navigation\cite{wang2025segnet4d, fan2020sne}. Recently, the cross-modal fusion of LiDAR point clouds and camera images has emerged as a promising solution for robust perception \cite{jiao2023lce}. A point-pixel fusion system combining LiDAR and camera can overcome the individual limitations of both sensors, producing reliable results that improve various tasks such as autonomous driving~\cite{guo2024lix}, SLAM~\cite{jiao2024real, liu2024edge, shi2024fast} and object recognition~\cite{kim2022dsqnet, fan2019pothole}. Point-pixel registration, which estimates the rigid transformation between point clouds and images, is a core component for efficient cross-modal fusion.

However, establishing accurate point-pixel correspondences remains highly challenging. Existing registration methods~\cite{he2025matchanything, bie2024image, lv2021lccnet, zhou2023differentiable, luo2025bevplace++} are often not suitable for sparse LiDAR scans, such as the 32-line sensor used in nuScenes, due to the limited number of observable points and the inherent modality gap between unstructured 3D point clouds and structured 2D images. Many of these methods rely on multi-frame accumulation or additional priors, which makes them impractical in scenarios such as target-less calibration for low-cost sensors or registration between sparse point clouds and images. This gap motivates the development of frameworks that can operate reliably under sparse, single-frame conditions. 
As illustrated in Fig.~\ref{fig:cover}, our method provides a unified framework for cross-modal registration, providing a robust solution for LiDAR-camera fusion in real-world robotic systems.

\begin{figure}
  \centering
  \includegraphics[width=1\linewidth]{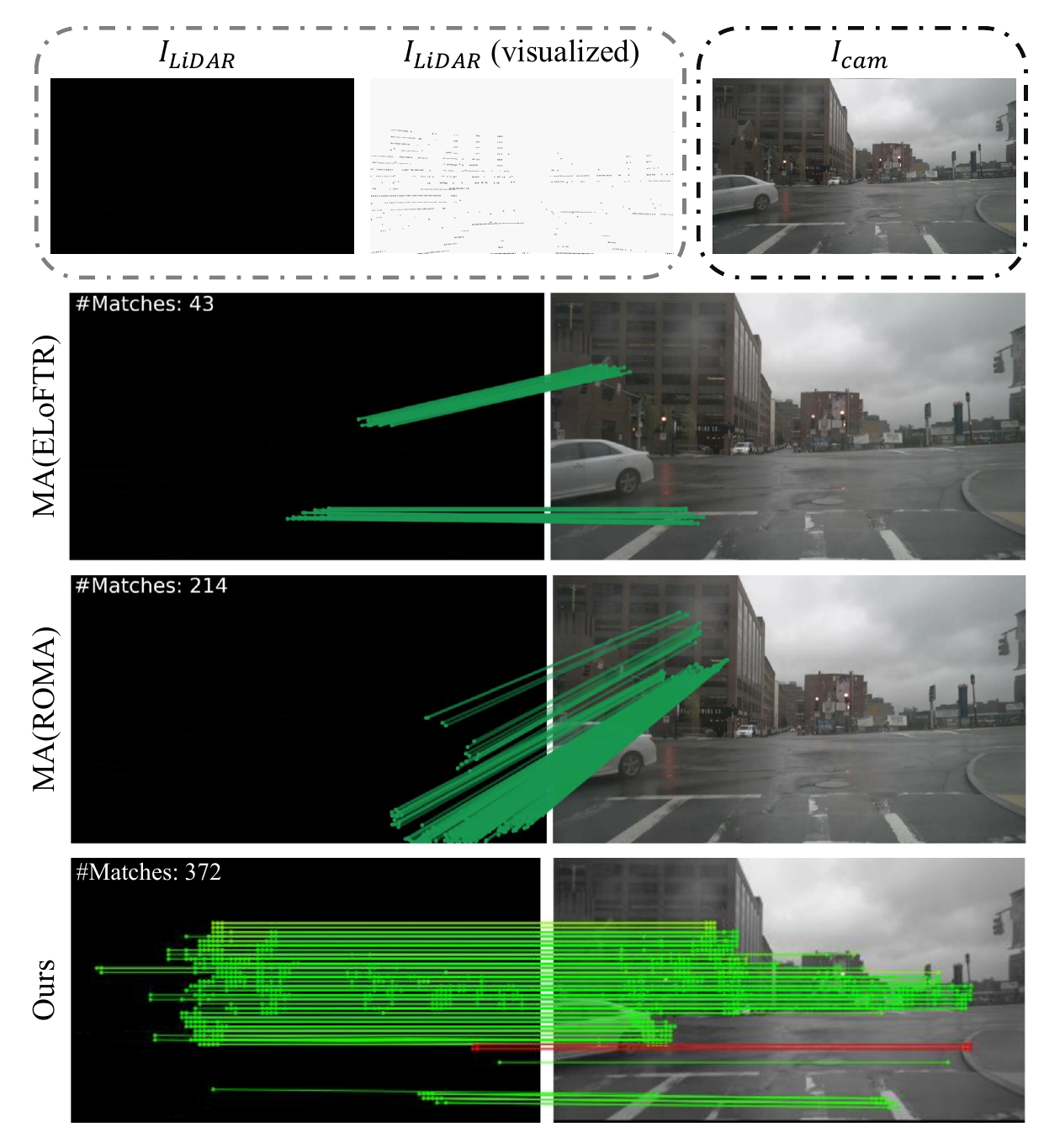}
  \caption{Visual comparison of correspondence estimation by MatchAnything (ELoFTR), MatchAnything (ROMA), and our proposed method.}
  \label{fig:vis_matching}
\end{figure}

Most existing works~\cite{schneider2017regnet, lv2021lccnet, iyer2018calibnet } only validate theirs approaches on KITTI datasets \cite{geiger2012we}, which focus on 64-line LiDARs and provide sufficiently point clouds for these methods to succeed. Research on 32-line sparse LiDARs is very limited. Even when 32-line LiDARs are studied~\cite{li2021deepi2p, bie2024image, zhou2023differentiable}, multi-frame accumulation is often required to obtain pseudo-dense point clouds, making it difficult to evaluate the methods on truly sparse single-frame scans. Methods that perform well on dense LiDARs, such as MatchAnything~\cite{he2025matchanything}, often fail under sparse conditions, as shown in Fig.~\ref{fig:vis_matching}. Consequently, for low-cost 32-line LiDARs, there are currently no methods that can simultaneously handle the modality gap and extreme sparsity. This motivates our work to develop approaches capable of establishing robust 3D-2D correspondences for sparse LiDAR data, enabling accurate calibration and registration in real-time and resource-constrained settings.

In this work, we revisit the classical projection-based approach and apply it directly to sparse, single-frame LiDAR data. Unlike prior arts that rely on dense reconstructions or depth completion, our approach only requires the intensity projections of the sparse point clouds.
Specifically, we render raw LiDAR points into a 2D view from the LiDAR perspective and extract features using standard 2D convolutional networks. 
Although this strategy works well with dense data, it confronts a huge performance degrade in cases where point clouds are sparse.
To address these shortcomings, we draw inspiration from detector-free correspondence approaches such as LoFTR~\cite{sun2021loftr}, which utilize attention mechanisms to reason about global context and establish dense correspondences without explicit keypoint detection. 
Building upon this insight, we adapt a detector-free, attention-based architecture to our cross-modal registration task. However, the standard Dual-Softmax matching strategy used in LoFTR is not well suited for the large modality gap and sparsity inherent to LiDAR projections, often resulting in ambiguous matches. To overcome this, we propose incorporating a repeatability score that acts as a soft visibility prior, reflecting the reliability of each projected point during matching. This allows the network to downweight unreliable or ambiguous regions such as occlusions, dynamic objects, and textureless surfaces, improving overall matching accuracy. As illustrated in Fig. \ref{fig:vis_matching}, our approach achieves high-quality 2D-2D correspondences between LiDAR projections and camera images on the nuScenes datasets, demonstrating robust and accurate matching even under challenging conditions such as sparsity and viewpoint variation. 
\begin{itemize}
    \item We revisit the classical projection-based framework and introduce a detector-free paradigm that enables reliable correspondence estimation between single-frame sparse LiDAR point clouds and camera images, which exhibits robust performance in sparse scenarios where previous methods fall short.
    \item We incorporate a repeatability-guided matching strategy that acts as a soft prior, enabling more reliable cross-modal correspondence under challenging conditions.
    \item We validate our approach on large-scale benchmarks, including KITTI Odometry, nuScenes and MIAS-LCEC-TF70, achieving state-of-the-art results. Especially on nuScenes, previous approaches typically rely on accumulated point clouds, whereas our approach achieves superior performance using only sparse point clouds.
\end{itemize}

\label{sec.outline}
The remainder of this article is structured as follows:
Sect. \ref{sec.relate_works} reviews state-of-the-art (SoTA) approaches of cross-modal registration.
Sect. \ref{sec.method} introduces our proposed algorithm. Sect. \ref{sec.experiment} presents experimental results and compares our method with SoTA methods.
Finally, in Sect. \ref{sec.conclusion}, we conclude this article and discuss potential future research directions.

\section{Related Work}
\label{sec.relate_works}
\subsection{Single-Modal Registration}

Single-modal registration aims to estimate transformations either between point clouds or between images of the same modality. For point clouds, point-point registration estimates a rigid transformation between two point sets by aligning their geometry. Beyond keypoint-based strategies, recent advances have demonstrated the effectiveness of keypoint-free and RANSAC-free pipelines. Approaches such as Predator~\cite{huang2021predator} and GeoTransformer~\cite{qin2022geometric} bypass traditional sampling and matching paradigms by leveraging attention mechanisms and global context, significantly improving robustness to partial overlaps and challenging initial poses.

For images, pixel-pixel registration aims to align images captured from different viewpoints or at different times by establishing correspondences directly between pixels \cite{han2024generalized, zhou2023e3cm}. Accurate registration underpins various applications, such as stereo camera extrinsic calibration \cite{zhao2024dive} and 3D reconstruction \cite{fan2018road}.Traditional registration pipelines often rely on handcrafted features such as SIFT~\cite{lowe2004distinctive} and ORB~\cite{rublee2011orb} to detect and match keypoints across images. However, these approaches often struggle under significant appearance variations, illumination changes, or low-texture scenes.
Learning-based approaches can be broadly divided into detector-based and detector-free paradigms. 
Detector-based methods~\cite{sarlin2020superglue, lindenberger2023lightglue} first detect salient keypoints and then extract descriptors for matching, achieving robustness under moderate viewpoint or illumination changes. 
More recently, detector-free approaches~\cite{sun2021loftr, wang2024efficient, edstedt2024roma} bypass explicit keypoint detection by directly predicting dense correspondences from image features, significantly improving performance in texture-less or repetitive regions.
Despite their effectiveness in single-modal scenarios, these approaches can not directly generalize to cross-modal registration, where LiDAR and camera data differ significantly in density, appearance, and noise characteristics.

\subsection{Cross-Model Registration}

Traditional point-pixel registration methods \cite{lv2015automatic, castorena2016autocalibration, yuan2021pixel, pandey2015automatic, tang2023robust} estimate the rigid transformation by aligning the cross-modal edges or mutual information (MI) extracted from LiDAR point cloud projections and camera images. While effective in specific scenarios with abundant features, these traditional methods heavily rely on well-distributed edges and rich texture, which largely compromise calibration robustness. Advances in deep learning techniques have driven significant exploration into enhancing traditional target-free algorithms. Some studies \cite{li2018automatic, ma2021crlf, wang2022automatic, han2021auto, liao2023se, zhu2020online, koide2023general, zhiwei2024lcec} explore attaching deep learning modules to their registration framework as useful tools to enhance calibration efficiency. For instance, \cite{ma2021crlf} accomplishes point-pixel registration by aligning road lanes and poles detected by semantic segmentation. Similarly, \cite{han2021auto} employs stop signs as calibration primitives and refines results over time using a Kalman filter. A recent study, DVL \cite{koide2023general}, introduced a novel point-based method that utilizes SuperGlue \cite{sarlin2020superglue} to establish direct 3D-2D correspondences between LiDAR and camera data. 
The study MIAS-LCEC \cite{zhiwei2024lcec}, developed a two-stage coarse-to-fine matching approach using the large vision model MobileSAM \cite{zhang2023faster} to improve cross-modal feature matching. While these method achieves high accuracy with dense point clouds when the sensors have overlapping fields of view, its performance significantly degrades in challenging scenarios with sparse or incomplete point clouds. In parallel, MatchAnything \cite{he2025matchanything} explores a detector-free paradigm for cross-modal matching and demonstrates impressive performance across various modalities. However, it also struggles to produce reliable correspondences when applied to sparse LiDAR projections, limiting its effectiveness in 3D-2D registration tasks.

To address these challenges, researchers attempt to use handcrafted keypoint detectors to enhance the quality of 3D-2D correspondences. For example, the study~\cite{feng20192d3d} extracts keypoints from images and point clouds, then learns feature descriptors with a triplet network to establish 3D-2D matches. However, the reliance on separate detectors for different modalities often leads to poor keypoint matching and low registration accuracy. In \cite{wang2021p2}, a unified network is utilized to simultaneously detect keypoints and extracts per-point descriptors in a single forward pass. They attempt to improve efficiency and matching accuracy by bridging the gap between image and point cloud modalities within an end-to-end framework. To avoid explicit keypoint matching, \cite{li2021deepi2p} formulates registration as a classification problem. It estimates camera pose by classifying 3D points into image frustums and refining the result with optimization. While this avoids correspondence search, the approach only provides coarse alignment due to its frustum-based approach.

Recent work focuses on learning dense 3D-2D correspondences. In \cite{ren2022corri2p}, pixel and point features are matched directly for fine registration. However, it struggles to learn discriminative cross-modal features, resulting in many incorrect matches. The study \cite{zhou2023differentiable} introduces voxels as an intermediate representation and uses a differentiable Perspective-\textit{n}-Point (PnP) solver for pose estimation. Still, these approaches often fail to build reliable correspondences due to the large modality gap between images and point clouds. 
Most recently, \cite{bie2024image} proposes a novel approach using virtual point clouds as bridges between modalities. Although this approach achieves state-of-the-art performance, its reliance on monocular depth estimation~\cite{lee2019big, zhang2024dcpi, feng2025vipocc, feng2024scipad} introduces fundamental limitations: (1) Monocular depth estimators inherently suffer from scale ambiguity and metric inaccuracy since they must infer 3D structure from single 2D images without absolute distance references; (2) The compounding errors from imperfect depth prediction directly propagate to the virtual point cloud generation, adversely affecting subsequent registration steps.

\begin{figure*}[t!]
    \centering
    \includegraphics[width=1\linewidth]{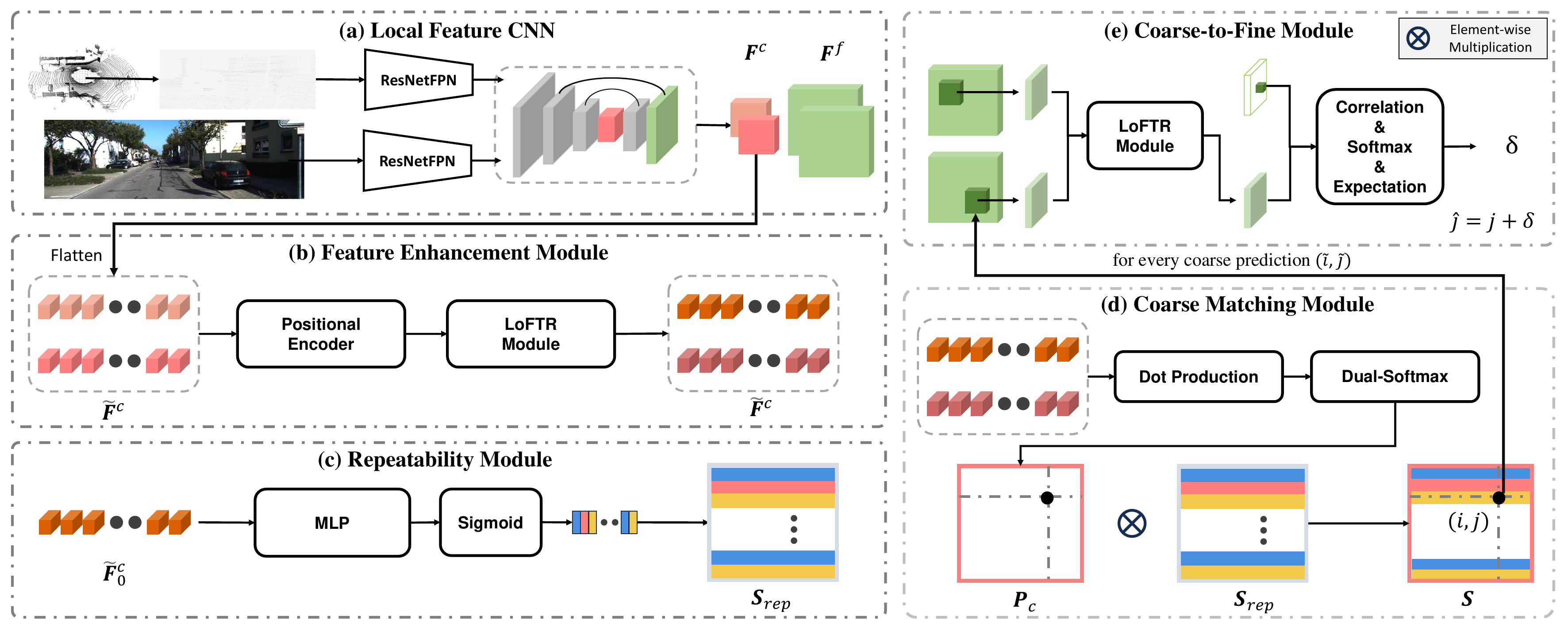}
    
    \caption{Architecture of the proposed multi-modal feature matching framework. Given a 4D point cloud and a real 2D RGB image \(\boldsymbol{I}\) as input,  
(a) the point cloud is projected onto the image plane to generate a virtual intensity image \(\boldsymbol{I}_{\text{LiDAR}}\) and a depth map \(\boldsymbol{D}_{\text{LiDAR}}\). The RGB image is converted into a grayscale image \(\boldsymbol{I}_{\text{cam}}\) for subsequent processing. A dual-path backbone extracts two levels of features from both \(\boldsymbol{I}_{\text{LiDAR}}\) and \(\boldsymbol{I}_{\text{cam}}\), producing a coarse-level feature map \(\boldsymbol{F}^c\) and a fine-level feature map \(\boldsymbol{F}^f\).
(b) The coarse-level features \(\boldsymbol{F}^c\) are flattened and enriched with positional encoding, resulting in \(\widetilde{\boldsymbol{F}}^c\), which is then processed by the LoFTR module. Multiple layers of self-attention and cross-attention are applied to enhance cross-modal consistency.
(c) Enhanced features are passed through a lightweight MLP, consisting of several linear layers and a sigmoid activation function, to regress repeatability scores \(\boldsymbol{S}_{rep}\), providing a per-pixel estimate of point repeatability. 
(d) At the coarse level, a similarity matrix \(\boldsymbol{S}_{\text{sim}}\) is computed using cosine similarity, followed by a Dual-Softmax operation to obtain the matching probability matrix \(\boldsymbol{P}_{c}\). The final score matrix is obtained by element-wise multiplication of the repeatability matrix and the probability matrix. 
(e) The coarse-to-fine module crops local windows around each coarse match from the fine-level features, applies a smaller transformer to compute correlation heatmaps, and estimates refined matches with sub-pixel accuracy.
}
\label{fig:framework}
\end{figure*}

\section{Methodology}
\label{sec.method}
Given a 4D point cloud with \((x, y, z, \text{intensity})\) and a real camera image \(\boldsymbol{I} \in \mathbb{R}^{H \times W \times 3}\), the goal is to estimate the rigid transformation \(\boldsymbol{T} = [\boldsymbol{R} \mid \boldsymbol{t}]\) between the LiDAR and camera coordinate systems. Here, \(\boldsymbol{R} \in SO(3)\) denotes the rotation matrix, and \(\boldsymbol{t} \in \mathbb{R}^3\) is the translation vector. 
In this section, the detailed methodology is presented, including preprocessing of inputs, local feature extraction, feature enhancement, repeatability scoring for keypoint repeatability, coarse-level matching, coarse-to-fine refinement of matches, and final pose estimation between the LiDAR and camera coordinate systems.

\begin{figure*}[t]
    \centering
    \begin{minipage}[t]{0.49\linewidth}
        \centering
        \includegraphics[width=\linewidth]{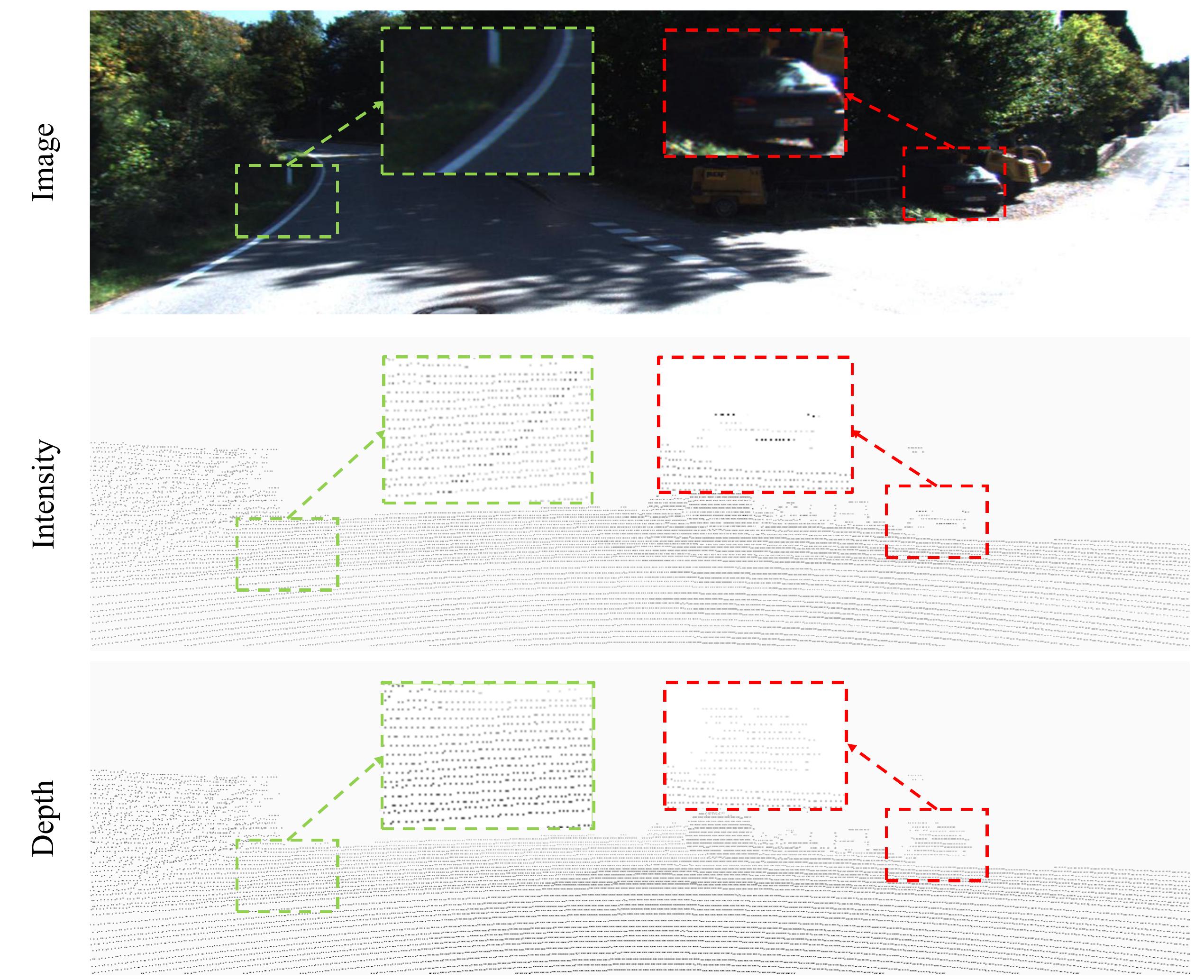}
        \caption{Visual comparison among the projected LiDAR intensity image, projected depth map, and the RGB image. In the depth map, colors encode the distance from the sensor. In contrast, the intensity image uses colors to represent the reflectivity of surfaces.}
        \label{fig:compare_kitti}
    \end{minipage}
    \hfill
    \begin{minipage}[t]{0.49\linewidth}
        \centering
        \includegraphics[width=\linewidth]{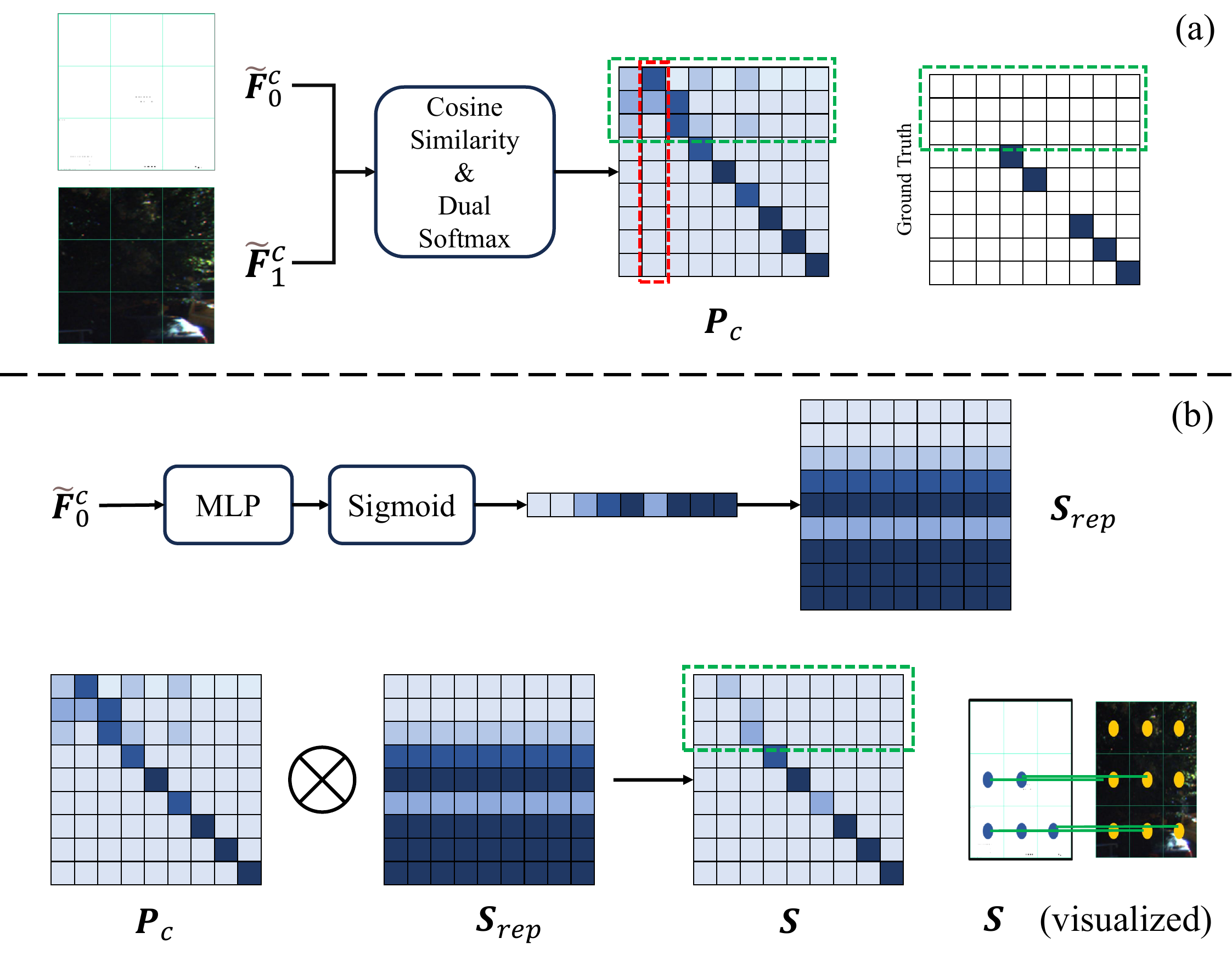}
        \caption{Dual-Softmax-only matching vs. repeatability scoring. Red boxes show mismatches not filtered by MNN. Green boxes highlight suppressed unreliable matches. Finally, we show the improved correspondence visualization.}
        \label{fig:repeat}
    \end{minipage}
\end{figure*}

\subsection{Feature Extraction and Enhancement}

As illustrated in the Fig. \ref{fig:framework}, given a 4D point cloud and an RGB image $\boldsymbol{I}$, our goal is to reduce the modality gap between the unstructured point cloud and the structured 2D image data. In our method, we first project the intensity from the raw point cloud, producing a virtual intensity (rather than depth) image $\boldsymbol{I}_{\text{LiDAR}} \in \mathbb{R}^{H \times W}$.
As depicted in Fig. \ref{fig:compare_kitti}, we provide a comparison between intensity projection, depth projection, and the RGB image. While depth maps are sensitive to the geometric continuity of objects, they often fail to capture fine-grained semantic or structural details, particularly in regions where depth remains constant. For example, lane markings and the road surface lie on the same plane and therefore exhibit little variation in the depth map. In contrast, these features are more distinguishable in the intensity projection, which captures significant differences in reflectance. Intensity is affected by material properties, making it more sensitive to surface characteristics. By leveraging intensity projections, we fine-tune local texture and material-related information, which improves the alignment between LiDAR and RGB modalities and enables more effective joint representation learning.

However, even after this transformation, some differences persist between the LiDAR intensity image and the grayscale image which must be explicitly addressed. The grayscale image is dense and continuous, offering rich textural and geometric cues that capture fine details such as edges and surface variations. In contrast, the intensity image derived from a single-frame point cloud remains inherently sparse. Moreover, reflectance intensity and RGB values correspond to fundamentally different physical properties, resulting in distinct response patterns across these modalities. Consequently, these inherent differences motivate the use of separate feature extractors tailored to each modality’s characteristics rather than forcing a shared representation prematurely. To this end, we adopt a ResNet-like~\cite{he2016deep} dual-backbone architecture with non-shared weights to extract hierarchical features from both $\boldsymbol{I}_{\text{LiDAR}}$ and $\boldsymbol{I}_{\text{cam}}$. Each branch independently processes one modality and outputs two feature maps, a coarse-level feature $\boldsymbol{F}^c \in \mathbb{R}^{\frac{H}{8} \times \frac{W}{8} \times C_c}$ and a fine-level feature $\boldsymbol{F}^f \in \mathbb{R}^{\frac{H}{2} \times \frac{W}{2} \times C_f}$. We denote the features from $\boldsymbol{I}_{\text{LiDAR}}$ as $\boldsymbol{F}^c_0$ and $\boldsymbol{F}^f_0$, and those from $\boldsymbol{I}_{\text{cam}}$ as $\boldsymbol{F}^c_1$ and $\boldsymbol{F}^f_1$.

Finally, the coarse-level features \(\boldsymbol{F}^c\) are first flattened and passed through a positional encoding layer, resulting in \(\widetilde{\boldsymbol{F}}^c \in \mathbb{R}^{(H/8 \cdot W/8) \times C_c}\), which are then fed into the LoFTR module~\cite{sun2021loftr}. The module employs multiple layers of self-attention and cross-attention to establish dense correlations between the two modalities, yielding a pair of enhanced features with consistent semantics across LiDAR and camera domains, facilitating more robust multi-modal fusion. 
Unlike prior works that either ignore intensity or treat it as a simple additional channel, our approach leverages intensity, instead of depth, as a modality-bridging cue within a detector-free registration pipeline. Through per-pixel training, intensity features and grayscale features are mapped into a shared feature space. In addition, sparse regions are suppressed to prevent isolated intensity values from degrading feature robustness and overall training effectiveness. This integration is central to our novelty, enabling robust point-pixel matching from a single sparse LiDAR scan without relying on multi-frame accumulation.

\subsection{Repeatability Scoring and Coarse Matching}

As shown in Fig. \ref{fig:repeat}, we assume an identity transformation between $\boldsymbol{I}_{\text{LiDAR}}$ and $\boldsymbol{I}_{\text{cam}}$. However, since $\boldsymbol{I}_{\text{LiDAR}}$ is generated by projecting sparse point clouds, some regions may suffer from missing projections, occlusions, or LiDAR noise, making depth consistency checks infeasible. As a result, valid correspondences are unavailable in certain areas. Fig. \ref{fig:repeat} (a) illustrates the probability matrix $\boldsymbol{P}_c$ derived from the $\boldsymbol{S}_\text{sim}$ via Dual-Softmax. Due to the normalization effect of Softmax, points in $\boldsymbol{I}_{\text{LiDAR}}$ without ground truth correspondences in $\boldsymbol{I}_{\text{cam}}$ are forced to exhibit uniform similarity across all positions, which may result in incorrect matches. Moreover, as shown by the red box, such mismatches cannot be effectively eliminated by mutual nearest neighbor (MNN) filtering, undermining both supervision and inference, especially in sparse projection scenarios. This issue is exacerbated by the inherent ``dense-near, sparse-far" nature of LiDAR data.

To address this issue, we introduce a lightweight repeatability scoring mechanism to explicitly evaluate the stability of LiDAR features, as illustrated in Fig. \ref{fig:repeat} (b). For each patch, a repeatability score is regressed to estimate its likelihood of being visible from another viewpoint. As shown in Eq.~\ref{equ:repeat}, each row of the similarity matrix is scaled by the corresponding visibility score, suppressing unreliable matches. This approach shares conceptual similarities with keypoint repeatability but differs in implementation. Unlike traditional repeatability, which predicts visibility from a single image across multiple views, our method leverages a cross-attention mechanism to infer visibility directly from the paired input image. The green box in Fig. \ref{fig:repeat} highlights this distinction during training. By incorporating repeatability scoring, we effectively alleviate the issue where the Softmax operator enforces uniform similarity in the absence of ground truth correspondences. Specifically, the enhanced coarse-level feature $\widetilde{\boldsymbol{F}}^c_0 $ passed through a lightweight multi-layer perceptron (MLP) followed by a Sigmoid activation $ \sigma(\cdot)$ to regress a per-pixel repeatability score:
\begin{equation}
\boldsymbol{S}_{\text{rep}} = \sigma(\text{MLP}(\widetilde{\boldsymbol{F}}^c_0)) \in \mathbb{R}^{(H/8 \cdot W/8)}.
\end{equation}

This produces a repeatability matrix that quantifies the expected stability and distinctiveness of each LiDAR feature.

To establish coarse-level correspondences, a similarity matrix $\boldsymbol{S}_{\text{sim}} \in \mathbb{R}^{(H/8 \cdot W/8) \times (H/8 \cdot W/8)}$ is computed between the flattened LiDAR and camera features using cosine similarity.
Then, a differentiable Dual-Softmax operator~\cite{rocco2018neighbourhood} produces a coarse matching probability matrix:
\begin{equation}
\boldsymbol{P}_c(i, j) = \text{Softmax}(\boldsymbol{S}_{\text{sim}}(i, \cdot))_j \cdot \text{Softmax}(\boldsymbol{S}_{\text{sim}}(\cdot, j))_i,
\end{equation}
where probability matrix $\boldsymbol{P}_c \in \mathbb{R}^{(H/8 \cdot W/8) \times (H/8 \cdot W/8)}$ reflects the confidence of correspondences between LiDAR and camera features.

It is noteworthy that the similarity matrix $\boldsymbol{S}_{\text{sim}}$ is computed using cosine similarity, which is widely adopted in traditional single-modal matching frameworks. Such similarity-based supervision implicitly encourages the descriptors from different modalities to be aligned in a shared feature space. Consequently, the training process promotes cross-modal consistency by making the descriptors at corresponding locations from the LiDAR and camera branches as similar as possible.

We further incorporate the repeatability scores into the confidence computation. The final confidence matrix is defined as follows:
\begin{equation}
\label{equ:repeat}
\boldsymbol{S}(i, j) = \boldsymbol{P}_c(i, j) \cdot \boldsymbol{S}_{\text{rep}}(i).
\end{equation}

To extract reliable coarse correspondences, we first apply the MNN criterion to the confidence matrix $\boldsymbol{P}_c$, effectively suppressing ambiguous and inconsistent matches. We then retain only those matches whose confidence exceeds a predefined threshold $\theta_c$, ensuring the quality of selected correspondences. This process yields a set of semantically consistent and geometrically plausible matches, which serve as a robust initialization for subsequent fine-level refinement:
\begin{equation}
\mathcal{M}_c = \{(i, j) \mid (i, j) \in \text{MNN}(\boldsymbol{S}),\ \boldsymbol{S}(i, j) \geq \theta_c\}.
\end{equation}

\subsection{Coarse-to-Fine Refinement}

Following~\cite{sun2021loftr}, a coarse-to-fine refinement strategy is employed to improve the localization accuracy of coarse correspondences. Each coarse match \((i, j)\) is refined within a local window on the fine-level feature maps using the LoFTR module, and the final sub-pixel accurate correspondences are obtained by selecting the peak in the local correlation heatmap. Specifically, given a set of coarse matches \(\mathcal{M}_c = \{(i, j)\}\), where \(i\) and \(j\) denote the indices in the downsampled coarse feature grids of \(I_{\text{LiDAR}}\) and \(I_{\text{cam}}\) respectively, each \((i, j)\) is mapped to its corresponding fine-level coordinate \(\hat{i}, \hat{j}\) in the higher-resolution feature map \(F_0^f\) and \(F_1^f\). Around \(\hat{i}\) and \(\hat{j}\), two local windows of size \(w \times w\) are cropped from the fine-level feature maps \(F_0^f\) and \(F_1^f\). These local features are then passed through a refinement transformer composed of several cross- and self-attention layers, which adaptively enhance local context and feature alignment.

The refined local features centered at \(\hat{i}\) and \(\hat{j}\) are then correlated: the feature vector at \(\hat{i}\) is treated as a query and is matched against all feature vectors within the window around \(\hat{j}\), yielding a correlation heatmap that encodes the matching probability distribution. To obtain the final refined position \(\hat{j}'\) on \(I_\text{{cam}}\), a soft-argmax operation is applied over the heatmap, resulting in a sub-pixel prediction. By repeating this process for all coarse matches, a set of fine-level correspondences \(\mathcal{M}_f\) is generated. These refined correspondences offer higher accuracy and are used as the output of the matching pipeline.

\subsection{Supervision}

The training objective combines coarse-level supervision, fine-level supervision, and a repeatability supervision:
\begin{equation}
L = L_c + L_f + L_r.
\end{equation}

For the coarse-level supervision, we supervise the predicted coarse confidence matrix $\boldsymbol{C}_c$ by minimizing a negative log-likelihood loss. Following the protocols of SuperGlue~\cite{sarlin2020superglue}, ground-truth matches $\mathcal{M}_c^{gt}$ are generated by projecting LiDAR points to the image plane using known camera poses and depth maps, followed by mutual nearest neighbor filtering with a reprojection error threshold. The loss is defined as follows:
\begin{equation}
    L_{c} = - \frac{1}{|\mathcal{M}_c^{gt}|} \sum_{(i, j) \in \mathcal{M}_c^{gt}} \log \boldsymbol{S}(i, j),
    \label{eq:loss_coarse}
\end{equation}
which not only encourages the network to assign high confidence to correct coarse correspondences, but also promotes the alignment of descriptors from the LiDAR intensity and grayscale image modalities in a shared feature space. By supervising cross-modal correspondences directly, the network learns to bridge modality gaps, facilitating more robust and semantically meaningful feature representations.

For the fine-level supervision, we adopt a weighted $\ell_2$ regression loss that leverages the uncertainty of the predicted local correlation heatmap. Given a refined match \((\hat{i}, \hat{j}')\), where \(\hat{j}'\) is predicted via soft-argmax and \(\hat{j}'_{gt}\) is its ground-truth projection, the loss is formulated as follows:
\begin{equation}
    L_{f} = \frac{1}{|\mathcal{M}_f|} \sum_{(\hat{i}, \hat{j}') \in \mathcal{M}_f} \frac{1}{\tau^2(\hat{i})} \left\| \hat{j}' - \hat{j}'_{gt} \right\|_2,
    \label{eq:loss_fine}
\end{equation}
where $\tau^2(\hat{i})$ represents the variance of the local heatmap and serves as a soft confidence weighting factor. 

For the repeatability supervision, a binary classification task is introduced to encourage physically consistent and modality-invariant feature extraction. 
To determine whether a patch in the LiDAR intensity image \( I_{\text{LiDAR}} \) has a valid and geometrically consistent correspondence in the grayscale image, a depth consistency check is performed. 
Specifically, for each patch, the top-left corner in \( I_{\text{LiDAR}} \) is back-projected into 3D space using the known depth map and LiDAR intrinsics. 
The resulting 3D point is then reprojected into the grayscale image using the camera extrinsics and intrinsics. 
If the projected point falls within image boundaries and the reprojected depth is consistent with the grayscale depth, the patch is considered to have a valid, physically consistent correspondence and is labeled as repeatable; otherwise, it is marked as non-repeatable. 
This process yields a binary ground-truth repeatability map \( \boldsymbol{S}_{\text{rep}}^{gt} \in \{0, 1\}^{H/8 \times W/8} \), which inherently excludes occluded or out-of-FOV regions. 
The predicted repeatability map \( \boldsymbol{S}_{\text{rep}} \in [0,1]^{H/8 \times W/8} \) is supervised using a binary cross-entropy loss, formulated as:
\begin{equation}
    L_{r} = \text{BCE}(\boldsymbol{S}_{\text{rep}}, \boldsymbol{S}_{\text{rep}}^{gt}),
    \label{eq:loss_rep}
\end{equation}
where BCE denotes the binary cross-entropy between prediction and ground truth. This supervision encourages the network to identify and emphasize geometrically reliable, modality-invariant regions across LiDAR and grayscale views.

\subsection{Pose Estimation}

With a set of refined correspondences between 3D LiDAR points and 2D image pixels, we estimate the extrinsic transformation \(\boldsymbol{T} = [\boldsymbol{R} \mid \boldsymbol{t}]\) using a PnP solver in conjunction with RANSAC~\cite{fischler1981random}. Each 3D-2D pair consists of a 3D point in the LiDAR coordinate system and a 2D projection in the camera image. Due to the sparsity of LiDAR data, especially in the periphery or occluded regions, the associated depth map may contain missing values. To mitigate this, we apply a nearest-neighbor filling strategy to recover missing depth values based on surrounding valid pixels. This ensures that all 2D matches used in pose estimation have corresponding depth values, thereby enabling robust lifting from image coordinates to 3D space.

\begin{table*}[t!]

\caption{Registration accuracy on the KITTI and nuScenes datasets. Lower is better for $e_t$ and $e_r$, higher is better for Acc. The best result is \textbf{bolded}. $^\dagger$Uses single-frame point clouds on nuScenes.}
\label{tab:kittiRegistration}

\centering
\fontsize{8}{9}\selectfont
\begin{tabular}{l|c|cc|c|cc|c}
\toprule
\multirow{2}*{Method} & \multirow{2}*{Type} 
& \multicolumn{3}{c|}{KITTI} 
& \multicolumn{3}{c}{nuScenes} \\
\cline{3-8}
& & $e_t$(m)$\downarrow$ & $e_r$($^\circ$)$\downarrow$ & Acc.$\uparrow$ & $e_t$(m)$\downarrow$ & $e_r$($^\circ$)$\downarrow$ & Acc.$\uparrow$ \\
\hline
\hline
vpc + CoFiNet \cite{yu2021cofinet}         & Point-to-Point & 5.74 $\pm$ 5.13 & 16.87 $\pm$ 23.36 & 16.04 & 3.77 $\pm$ 6.31 & 12.29 $\pm$ 11.48 & 2.53 \\
vpc + D3Feat \cite{bai2020d3feat}           & Point-to-Point & 6.61 $\pm$ 5.28 & 9.79 $\pm$ 17.74  & 32.48 & 3.05 $\pm$ 3.68 & 9.16 $\pm$ 8.07   & 18.92 \\
vpc + PREDATOR \cite{huang2021predator}       & Point-to-Point & 4.12 $\pm$ 12.81 & 7.67 $\pm$ 13.67 & 48.65 & 2.63 $\pm$ 2.40 & 4.78 $\pm$ 4.03  & 38.65 \\
vpc + GeoTransformer \cite{qin2023geotransformer} & Point-to-Point & 3.43 $\pm$ 5.86 & 6.15 $\pm$ 11.37 & 70.34 & 2.08 $\pm$ 2.75 & 4.30 $\pm$ 3.41 & 42.87 \\
\hline
Grid Cls. + PnP \cite{li2021deepi2p} & Image-to-Point & 3.64 $\pm$ 3.46 & 19.19 $\pm$ 28.96 & 11.22 & 3.02 $\pm$ 2.40 & 12.66 $\pm$ 21.01 & 2.45 \\
DeepI2P (3D) \cite{li2021deepi2p}     & Image-to-Point & 4.06 $\pm$ 3.54 & 24.73 $\pm$ 31.69 & 3.77  & 2.88 $\pm$ 2.12 & 20.65 $\pm$ 12.24 & 2.26 \\
DeepI2P (2D) \cite{li2021deepi2p}     & Image-to-Point & 3.59 $\pm$ 3.21 & 11.66 $\pm$ 18.16 & 25.95 & 2.78 $\pm$ 1.99 & 4.80 $\pm$ 6.21   & 38.10 \\
CorrI2P \cite{ren2022corri2p}         & Image-to-Point & 3.78 $\pm$ 65.16 & 5.89 $\pm$ 20.34 & 72.42 & 3.04 $\pm$ 60.76 & 3.73 $\pm$ 9.03  & 49.00 \\
VP2P \cite{zhou2023differentiable}           & Image-to-Point & 0.75 $\pm$ 1.13 & 3.29 $\pm$ 7.99 & 83.04 & 0.89 $\pm$ 1.44 & 2.15 $\pm$ 7.03 & 88.33 \\
Bie \etal \cite{bie2024image} & Image-to-Point & 0.61 $\pm$ 0.93 & 2.89 $\pm$ 3.21 & 86.72 & 0.67 $\pm$ 0.87 & 1.84 $\pm$ 3.41 & \textbf{89.47} \\

\hline
LoFTR(fine-tuned)$^\dagger$ + EPnP \cite{sun2021loftr}                            & Image-to-Image & 0.40 $\pm$ 1.42 & 1.88 $\pm$ 11.70 & 97.68 & 0.91 $\pm$ 0.96 & 3.37 $\pm$ 3.87 & 66.00 \\
Ours$^\dagger$ + EPnP                        & Image-to-Image & \textbf{0.25} $\pm$ \textbf{0.67} & \textbf{0.86} $\pm$ \textbf{1.89} & \textbf{98.75} & \textbf{0.65} $\pm$ \textbf{0.76} & \textbf{1.73} $\pm$ \textbf{2.61} & 84.33 \\
\bottomrule 
\end{tabular}
\end{table*}

\section{Experiment}
\label{sec.experiment}
\subsection{Dataset}

We evaluate our approach on the task of point-pixel registration using two widely adopted autonomous driving benchmarks: KITTI Odometry~\cite{geiger2012we} and nuScenes~\cite{caesar2020nuscenes}. Both datasets contain synchronized image and LiDAR point cloud data collected by 2D cameras and 4D LiDAR sensors.

\subsubsection{KITTI Odometry} 
Following~\cite{zhou2023differentiable, bie2024image}, we use sequences 00-08 for training and 09-10 for testing. To compare with calibration methods~\cite{ma2021crlf, zhiwei2024lcec}, we adopt sequence 02 for training and the rest for testing, ensuring both fairness and diversity. Mis-registration LiDAR projections are generated by applying +1 m translation on the ground plane and \(\pm10^\circ\) rotation around the up-axis. Images are resized to 840 pixels on the longer side, and single-frame LiDAR data is used without accumulation. We subsample by selecting one frame every 4 frames for training and every 5 frames for testing.

\subsubsection{nuScenes}
We follow the official split of the nuScenes dataset, using 850 scenes for training and 150 scenes for testing. Compared to KITTI, nuScenes is equipped with a 32-line LiDAR, which is significantly sparser than the 64-line LiDAR used in KITTI, making the registration task more challenging. Prior works such as~\cite{zhou2023differentiable, bie2024image} rely on aggregating multiple sweeps to accumulate denser point clouds. In contrast, we demonstrate that our approach achieves competitive performance using only single-frame LiDAR point clouds without any accumulation. The image-point cloud pairs are generated using the official nuScenes SDK, where each image is taken from the current frame, and the corresponding point cloud is  projected into an image plane within a mis-registration range originated from the front camera. To ensure consistency in input resolution, we resize each image such that its longer side is scaled to 840 pixels while maintaining the original aspect ratio. Similar to the processing of KITTI, we uniformly sample 500 image-point cloud pairs from the training scenes and 300 pairs from the testing scenes to construct our training and evaluation sets. This uniform sampling strategy helps balance scene diversity while avoiding redundancy across highly similar consecutive frames.

\begin{table*}[t!]
\caption{Quantitative comparisons with SoTA target-free LCEC approaches on the 00 sequence of KITTI odometry. The best result is \textbf{bolded}. The upper part shows methods that directly regress poses from image pairs, while the lower part includes methods that first perform matching and then solve for the pose.}
\centering
\fontsize{8}{9}\selectfont
\setlength{\tabcolsep}{15pt}
\begin{tabular}{l|c|c@{\hspace{0.15cm}}c|c@{\hspace{0.15cm}}c@{\hspace{0.15cm}}c|c@{\hspace{0.15cm}}c@{\hspace{0.15cm}}c}
\toprule
\multirow{2}*{Approach} & \multirow{2}*{Initial Range} & \multicolumn{2}{c|}{Magnitude}
& \multicolumn{3}{c|}{Rotation Error ($^\circ$)} & \multicolumn{3}{c}{Translation Error (m)} \\
\cline{3-10}
& & $e_r$ ($^\circ$) & $e_t$ (m) & Yaw & Pitch & Roll  & X & Y & Z \\
\hline
\hline
CalibRCNN \cite{shi2020calibrcnn} &$\pm10^\circ / \pm 0.25m$ &0.805 &\textbf{0.093}	&0.446	&0.640	&0.199	&0.062	&0.043 &0.054 \\
CalibDNN \cite{zhao2021calibdnn} &$\pm10^\circ / \pm 0.25m$ &1.021 &0.115	&0.200	&0.990	&0.150	&0.055	&0.032	&0.096 \\
RegNet \cite{schneider2017regnet} &$\pm20^\circ / \pm 1.5m$ &0.500 &0.108	&0.240	&\textbf{0.250}	&0.360	&0.070	&0.070	&0.040 \\
LCCNet \cite{lv2021lccnet} &$\pm10^\circ / \pm 1.0m$ &1.418 &0.600  &0.455 &0.835 &0.768 &0.237 &0.333  &0.329 \\
RGGNet \cite{yuan2020rggnet} &$\pm20^\circ / \pm 0.3m$ &1.29 &0.114	&0.640 &0.740&0.350 &0.081 &\textbf{0.028} &\textbf{0.040} \\
CalibNet \cite{iyer2018calibnet} &$\pm10^\circ / \pm 0.2m$ &5.842 &0.140	&2.873 &2.874 &3.185 &0.065 &0.064 &0.083 \\
\hline
Borer \etal \cite{borer2024chaos}&$\pm1^\circ / \pm 0.25m$  &\textbf{0.455} &0.095	&0.100 &0.440 &\textbf{0.060}  &\textbf{0.037} &0.030 &0.082 \\
CRLF \cite{ma2021crlf} &- &0.629	&4.118	&\textbf{0.033}	&0.464	&0.416	&3.648	&1.483	&0.550 \\
UMich \cite{pandey2015automatic} &- &4.161	&0.321	&0.113	&3.111	&2.138	&0.286	&0.077	&0.086 \\
HKU-Mars \cite{yuan2021pixel} &- &33.84	&6.355	&19.89	&18.71	&19.32	&3.353	&3.232	&2.419 \\
DVL \cite{koide2023general}  &-&122.1 &5.129	&48.64	&87.29	&98.15	&2.832	&2.920	&1.881 \\
MIAS-LCEC \cite{zhiwei2024lcec}  &-&5.385	&1.014	&1.574	&4.029	&4.338	&0.724	&0.383	&0.343 \\
Ours + EPnP   &- &0.529	&0.150	&0.108	&0.464	&0.121	&0.096	&0.032	&0.084 \\
\bottomrule
\end{tabular}
\label{tab.cmp_kitti00}
\end{table*}

\begin{table*}[t!]
\caption{Quantitative comparisons with SoTA LCEC approaches on KITTI odometry (01-08 sequences). The best result is \textbf{bolded}.}
\centering
\fontsize{8}{9}\selectfont
\begin{tabular}{l
|c@{\hspace{0.15cm}}c
|c@{\hspace{0.15cm}}c
|c@{\hspace{0.15cm}}c
|c@{\hspace{0.15cm}}c
|c@{\hspace{0.15cm}}c
|c@{\hspace{0.15cm}}c
|c@{\hspace{0.15cm}}c}
\toprule
\multirow{2}*{Approach} 
& \multicolumn{2}{c|}{01} 
& \multicolumn{2}{c|}{03}  
& \multicolumn{2}{c|}{04} 
& \multicolumn{2}{c|}{05} 
& \multicolumn{2}{c|}{06}
& \multicolumn{2}{c|}{07} 
& \multicolumn{2}{c}{08} \\
\cline{2-15}
& $e_t(m)$ & $e_r(^\circ)$
& $e_t(m)$ & $e_r(^\circ)$  
& $e_t(m)$ & $e_r(^\circ)$ 
& $e_t(m)$ & $e_r(^\circ)$ 
& $e_t(m)$ & $e_r(^\circ)$ 
& $e_t(m)$  & $e_r(^\circ)$ 
& $e_t(m)$  & $e_r(^\circ)$  \\ 
\hline
\hline
CRLF \cite{ma2021crlf}  
&7.363 &0.623
&6.007 &\textbf{0.845}
&0.372 &\textbf{0.601}
&5.961 &\textbf{0.616}
&25.762 &\textbf{0.615}
&1.807 & \textbf{0.606}
&5.376 &\textbf{0.625} \\
UMich \cite{pandey2015automatic}  
&0.305 &2.196
&\textbf{0.316} &3.201
&0.348 &2.086
&0.356 &3.526
&0.353 &2.914
&0.368 &3.928
&0.367 &3.722 \\
HKU-Mars \cite{yuan2021pixel}  
&3.770 &20.73
&3.493 &21.99
&0.965 &4.943
&6.505 &34.42
&7.437 &25.20
&7.339 &33.10
&8.767 &26.62 \\
DVL \cite{koide2023general}  
&2.514 &112.0
&4.711 &124.7
&4.871 &113.5
&4.286 &123.9
&5.408 &128.9
&5.279 &124.7
&4.461 &126.2 \\
MIAS-LCEC \cite{zhiwei2024lcec}  
&\textbf{0.300} &\textbf{0.621}
&0.324 &1.140
&0.369 &0.816
&0.775 &4.768
&0.534 &2.685
&1.344 &11.80
&0.806 &5.220 \\
Ours + EPnP
&0.344 &1.065
&0.384 &1.838
&\textbf{0.267} &0.821
&\textbf{0.219} &0.771
&\textbf{0.191} &0.764
&\textbf{0.155} &\textbf{0.606}
&\textbf{0.200} &0.725 \\
\bottomrule
\end{tabular}
\label{tab.rescmp_kitti_01_08}
\end{table*}

\subsubsection{MIAS-LCEC-TF70}
MIAS-LCEC-TF70~\cite{zhiwei2024lcec} is a challenging dataset that contains 60 pairs of 4D dense point clouds (including spatial coordinates with intensity data) and RGB images, collected using a Livox Mid-70 LiDAR and a MindVision SUA202GC camera, from a variety of indoor and outdoor environments, under various scenarios as well as different weather and illumination conditions. This dataset is divided into six subsets: residential community, urban freeway, building, challenging weather, indoor, and challenging illumination. However, due to the limited size of the dataset and our focus on outdoor scenes, we use only the outdoor subsets and randomly split them evenly into training and testing sets.

\subsection{Baseline and Metrics}
\subsubsection{Baseline} 
We compare our approach with representative registration methods from four categories: (1) Point-point registration methods that align 3D point clouds, e.g., CoFiNet~\cite{yu2021cofinet}; (2) Image-point registration methods that match 2D image features to 3D points, e.g., VP2P~\cite{zhou2023differentiable}; (3) Image-image registration methods that align features across images, including the fine-tuned LoFTR~\cite{sun2021loftr}; (4) LiDAR-camera extrinsic calibration (LCEC) approaches, including both regression-based methods (directly regressing 6-DoF poses via end-to-end deep learning, e.g., CalibNet~\cite{iyer2018calibnet}) and matching-based methods (estimating extrinsic parameters from cross-modal feature correspondences, e.g., DVL~\cite{koide2023general}).

\label{sec:exp.metrics}
\subsubsection{Metrics} Following the practice of~\cite{li2021deepi2p, zhou2023differentiable, ren2022corri2p, bie2024image}, we adopt the Relative Rotation Error ($e_r$) and Relative Translation Error ($e_t$) as our main evaluation metrics, measured in degrees and meters, respectively. To ensure fair comparison, we follow the evaluation protocol of~\cite{zhou2023differentiable}, which computes the average $e_r$ and $e_t$ over all samples, without discarding any samples with large registration errors. Each sample consists of a point cloud from a single LiDAR frame and an image captured by a camera at the corresponding time. This differs from the CorrI2P~\cite{ren2022corri2p} setting, where samples with large errors are removed prior to averaging. Additionally, we report the registration accuracy (Acc.), which measures the proportion of successful fine registrations whose $e_r$ and $e_t$ are below predefined thresholds of 5$^\circ$ and 2 meters, respectively. These thresholds are commonly used in prior works~\cite{zhou2023differentiable, bie2024image, ren2022corri2p} to indicate high-quality registration. Although we do not explicitly discard samples with large registration errors, samples that fail to produce a sufficient number of correspondences (fewer than 4) in extremely challenging scenes are excluded from the average error computation, as EPnP-based pose estimation cannot be performed. However, such cases are still counted as failures when computing registration accuracy. Furthermore, we also report the matching precision following \cite{sarlin2020superglue}, which measures the proportion of predicted correspondences that satisfy the epipolar geometry constraint.

\subsection{Implementation Details}
We initialize our model with LoFTR pretrained weights and train on KITTI sequence 02 for 10 epochs. For fair comparison, we additionally fine-tune for 5 epochs on sequences 00-08. On nuScenes, the model is fine-tuned using KITTI-pretrained weights, and further adapted to MIAS-LCEC-TF70 for 10 epochs. The learning rate is $1 \times 10^{-4}$ for KITTI pretraining and $1 \times 10^{-5}$ for fine-tuning. All experiments are conducted on an NVIDIA RTX 3090 GPU with batch size 1.

\subsection{Registration Accuracy}

The registration results on the KITTI and nuScenes dataset are presented in Table~\ref{tab:kittiRegistration}. Our proposed approach significantly outperforms all existing approaches in both translation and rotation accuracy. For instance, compared with bie \etal~\cite{bie2024image}, the previous SoTA method, our approach reduces the translation error from 0.61\,m to 0.25\,m and the rotation error from 2.89$^\circ$ to 0.86$^\circ$. Additionally, we boost the registration accuracy from 86.72\% to 98.75\%, setting a new state-of-the-art on the KITTI benchmark. This improvement stems from our detector-free matching paradigm, enhanced by an attention-based architecture, which effectively overcomes the inherent challenges in traditional projection-based frameworks. By addressing the core difficulties of projection-based matching, our method enables a re-examination of projection as a viable and effective design choice. The projection operation naturally transforms unstructured point clouds into structured representations, which aligns well with the grid-based nature of images. This structured alignment facilitates bridging the modality gap between point clouds and images. When combined with our attention-driven matching strategy, this design leads to significant improvements in both robustness and accuracy for cross-modality registration tasks. 

Compared with LoFTR~\cite{sun2021loftr}, a detector-free image matching baseline fine-tuned for registration, our method still achieves superior performance, reducing the translation and rotation errors from 0.40\,m and 1.88$^\circ$ to 0.25\,m and 0.86$^\circ$, respectively, while improving the accuracy from 97.68\% to 98.75\%. It is also worth noting that 0.54\% of the samples in LoFTR's predictions failed to yield more than four correspondences and were consequently discarded, whereas our method exhibited a lower failure rate of only 0.18\%, demonstrating higher robustness in challenging registration scenarios. This improvement can be attributed to the introduction of the repeatability scoring mechanism in our framework. Unlike LoFTR, which treats all pixel-level features equally in matching, our method explicitly assigns a repeatability score to each projected point from the LiDAR modality. This score reflects the likelihood that a point remains consistent under viewpoint changes, effectively filtering out unstable or occlusion-prone regions. By incorporating this mechanism, we better mitigate the issues caused by perspective distortion, occlusions, and ambiguous associations in the Dual-Softmax matching process. As a result, our method achieves more reliable and accurate correspondences, especially under sparse or geometrically complex conditions.

Similarly, on the nuScenes dataset, our method consistently achieves superior registration performance. Compared with bie \etal~\cite{bie2024image}, our approach reduces the translation error from 0.67\,m to 0.65\,m and the rotation error from 1.84$^\circ$ to 1.73$^\circ$. It is important to note that, unlike prior methods such as bie \etal, which leverage accumulated point clouds to mitigate sparsity and enhance registration robustness, our method operates on single-frame LiDAR data without any temporal accumulation. This improvement is mainly due to our detector-free matching paradigm and the projection-based framework. Compared with LoFTR~\cite{sun2021loftr}, our method reduces the translation error from 0.91\,m to 0.65\,m and the rotation error from 3.37$^\circ$ to 1.73$^\circ$, and significantly improves the accuracy from 66.00\% to 84.33\%. This improvement is attributed to our use of an additional backbone for cross-modality representation learning and a repeatability scoring mechanism that enhances matching reliability. The difference performance between KITTI and nuScenes is relatively small for previous methods mainly because they leverage ground-truth poses to accumulate multi-frame point clouds. However, we argue that such accumulation is impractical in real-world applications, as it relies on access to accurate poses beforehand. In contrast, our method only requires single-frame LiDAR data without temporal accumulation. Although this leads to a performance drop on the nuScenes dataset due to the sparser point clouds and more challenging scenes, our approach still outperforms previous methods, demonstrating its strong generalization ability and robustness in real-world scenarios.

\begin{table*}[t!]
\caption{Quantitative comparisons with SoTA LCEC approaches on MIAS-LCEC-TF70. The best result is \textbf{bolded}.}
\centering
\fontsize{8}{9}\selectfont
\renewcommand{\arraystretch}{1.2}
\begin{tabular}{
    l|
    >{\centering\arraybackslash}p{0.8cm}
    >{\centering\arraybackslash}p{0.8cm}|
    >{\centering\arraybackslash}p{0.8cm}
    >{\centering\arraybackslash}p{0.8cm}|
    >{\centering\arraybackslash}p{0.8cm}
    >{\centering\arraybackslash}p{0.8cm}|
    >{\centering\arraybackslash}p{0.8cm}
    >{\centering\arraybackslash}p{0.8cm}|
    >{\centering\arraybackslash}p{0.8cm}
    >{\centering\arraybackslash}p{0.8cm}
}
\toprule
\multirow{2}*{Approach} & \multicolumn{2}{c|}{\makecell{Residential\\Community}} & \multicolumn{2}{c|}{\makecell{Urban\\Freeway}} & \multicolumn{2}{c|}{Building} & \multicolumn{2}{c|}{\makecell{Challenging\\Weather}} & \multicolumn{2}{c}{All} \\
\cline{2-11}
& $e_t(m)$ & $e_r(^\circ)$ & $e_t(m)$ & $e_r(^\circ)$ & $e_t(m)$ & $e_r(^\circ)$ & $e_t(m)$ & $e_r(^\circ)$ & $e_t(m)$ & $e_r(^\circ)$ \\

\hline
\hline
CRLF \cite{ma2021crlf}        & 0.464 & 1.594 & 0.140 & 1.582 & 20.17 & 1.499 & 2.055 & 1.646 & 5.707 & 1.580 \\
UMich \cite{pandey2015automatic} & 0.387 & 4.829 & 0.166 & 2.267 & 0.781 & 11.914 & 0.310 & 1.851 & 0.411 & 5.215 \\
HKU-Mars \cite{yuan2021pixel} & 1.208 & 2.695 & 1.956 & 2.399 & 0.706 & 1.814 & 1.086 & 2.578 & 1.239 & 2.372 \\
DVL \cite{koide2023general}   & 0.063 & 0.193 & 0.124 & 0.298 & 0.087 & 0.200 & 0.052 & 0.181 & 0.082 & 0.218 \\
MIAS-LCEC \cite{zhiwei2024lcec} & \textbf{0.050} & 0.190 & \textbf{0.111} & \textbf{0.291} & \textbf{0.072} & 0.198 & \textbf{0.046} & \textbf{0.177} & \textbf{0.070} & \textbf{0.214} \\
Ours + EPnP                         & 0.088 & \textbf{0.151} & 0.167 & 0.304 & 0.227 & \textbf{0.197} & 0.125 & 0.304 & 0.152 & 0.239 \\


\bottomrule
\end{tabular}
\label{tab.lcec_cal}
\end{table*}

\subsection{Calibration Accuracy}
\subsubsection{Sparse Point Cloud}
Table~\ref{tab.cmp_kitti00} presents quantitative comparisons between our method and a series of state-of-the-art target-free LCEC approaches on the KITTI Odometry sequence 00. The upper part of the table includes methods that directly regress poses from image pairs, while the lower part lists methods that first perform cross-modal matching and then estimate poses. 
According to the experimental results, our method significantly outperforms most matching-based LCEC approaches, while performing slightly worse than some of the regression-based methods. Note that we only train on sequence 02, while most direct regression methods are trained on more KITTI sequences. Although these regression methods achieve strong performance in some metrics, they rely on black-box regression without explicit geometric constraints and require large training data, making them sensitive to modality gaps, viewpoint changes, and dynamics.
In contrast, our method first establishes reliable correspondences between LiDAR intensity and RGB images, and then estimates the pose via EPnP. This design offers better interpretability and robustness, while also achieving competitive results with regression-based methods and outperforming all existing matching-based methods.
On the other hand, compared to previous matching-based methods, our method significantly outperforms them, reducing the rotation error by 5.385$^\circ$ and the translation error by 0.864\,m. This improvement benefits from the introduction of an enhanced detector-free matching paradigm, which leverages supervised learning to establish robust correspondences between sparse point clouds and images.

In addition, we compare our approach with several matching-based LCEC methods on the KITTI odometry benchmark (sequences 01-08), as shown in Table~\ref{tab.rescmp_kitti_01_08}. 
Our method achieves the lowest average translation error (0.252 m) and second-lowest rotation error (0.941$^\circ$), ranking first or second in almost all sequences. Compared to MIAS-LCEC~\cite{zhiwei2024lcec}, which performs well in textured scenes, our method shows more balanced performance across diverse environments.
Traditional methods such as CRLF~\cite{ma2021crlf}, UMich~\cite{pandey2015automatic}, and DVL~\cite{koide2023general} suffer from either large translation or rotation errors due to reliance on sparse features, semantic priors, or flow consistency. Our advantage comes from explicitly learning dense, geometry-aware features across modalities. This allows accurate correspondence supervision without depending on external cues or handcrafted steps, highlighting the effectiveness of our cross-modal matching paradigm for LiDAR-camera calibration. However, our approach shows a relative weakness on sequence 01, which primarily consists of highway scenes where LiDAR points mostly fall on the road surface. This results in fewer distinctive and easily matchable road signs, limiting correspondence quality. Additionally, the performance on sequence 03 is somewhat degraded compared to sequence 02, likely due to a significant data distribution shift between these sequences that affects matching accuracy.

\subsubsection{Dense Point Cloud}
As shown in Table~\ref{tab.lcec_cal}, we compare our approach with several SoTA LCEC methods on the MIAS-LCEC-TF70~\cite{zhiwei2024lcec} benchmark, including CRLF~\cite{ma2021crlf}, UMich~\cite{pandey2015automatic}, HKU-Mars~\cite{yuan2021pixel}, DVL~\cite{koide2023general}, and MIAS-LCEC~\cite{zhiwei2024lcec}. The metrics used are the translation error and rotation error, evaluated across four scene categories: Residential Community, Urban Freeway, Building, and Challenging Weather, as well as the overall performance.

Our method achieves competitive results, with the best rotation error in the Residential Community and Building scenes, and a strong overall rotation performance ($e_r = 0.239$), which is close to the state-of-the-art. However, the translation error in the Urban Freeway and Building scenes is relatively higher compared to other methods. We attribute the lower performance in these two scenes to the limited number of available training samples. Specifically, the Urban Freeway and Building sequences contain only 6 and 7 pairs respectively, resulting in merely 3 pairs available for fine-tuning after dataset splitting. The small amount of training data significantly constrains the network's ability to adapt to these scenes, especially under the domain-specific geometric and environmental variations.

\subsection{Computational Cost and Runtime Analysis}
For preprocessing, the total computational cost mainly depends on the LiDAR point density. On an AMD Ryzen 5 5600G processor, this step takes approximately 3.5 ms for 40 k points, 16 ms for 200 k points, and 38 ms for 500 k points. During inference, both runtime and memory usage vary with the projection resolution. At a dense projection resolution of $840 \times 560$, inference requires about 138 ms per pair and 4,560 MB of GPU memory, whereas a sparse projection of $840 \times 472$ takes approximately 127 ms and 2,689 MB.

\begin{table}[t!]
\centering
\fontsize{8}{9}\selectfont
\caption{Ablation study of different modules. The upper part reports results on the KITTI 09 sequence, and the lower part shows results on the nuScenes test dataset.}
\label{tab:ablation}
\begin{tabular}{lccc}
\toprule
Module & Precision & $e_t$ & $e_r$ \\
\midrule
Baseline (fine-tuned) & \textbf{96.00} & 0.32 & 0.81 \\
w/o Dual Backbone & 95.38 & \textbf{0.22} & \textbf{0.71} \\
w/o Pepeat. Socre & 95.40 & 0.30 & 0.75 \\
Full & 95.20 & \textbf{0.22} & 0.74 \\
\midrule
Baseline (fine-tuned) & 85.93 & 0.91 & 4.37 \\
w/o Dual Backbone & 84.49 & 0.89 & 3.49 \\
w/o Pepeat. Socre & 88.45 & \textbf{0.65} & 2.03 \\
Full & \textbf{88.75} & \textbf{0.65} & \textbf{1.73} \\
\bottomrule
\end{tabular}
\end{table}

\subsection{Ablation Study}
Table~\ref{tab:ablation} summarizes the ablation study on KITTI and nuScenes. On KITTI, a single shared backbone already yields strong performance (precision: 96.00, $e_t = 0.32$, $e_r = 0.81$), as the 64-line LiDAR retains sufficient structural cues. In contrast, the sparser 32-line LiDAR in nuScenes causes a significant drop (precision: 85.93, $e_t = 0.91$, $e_r = 3.37$), making the dual backbone more beneficial. Although the gain on KITTI is marginal, it provides a stronger foundation on nuScenes.

The repeatability scoring further improves robustness on sparse data. Removing it increases rotation error from 1.73 to 2.03 and slightly reduces precision, showing its role in filtering unstable regions. Finally, combining both designs achieves the SoTA overall performance on nuScenes (precision: 88.75, $e_t = 0.65$, $e_r = 1.73$), demonstrating their complementary effects in enhancing cross-modal registration under sparse LiDAR conditions.

\section{Conclusion}
\label{sec.conclusion}

In this paper, we extend the classical projection-based framework for point-pixel registration by incorporating attention-based matching and a repeatability-guided prior, enabling robust performance with sparse, single-frame LiDAR data.
By projecting LiDAR intensity and leveraging attention-based dense matching, our approach effectively bridges the modality gap between 3D point clouds and 2D images. To further enhance reliability, we introduce a repeatability-guided score that acts as a soft prior, enabling the suppression of unreliable correspondences caused by occlusion and textureless regions. Extensive experiments on KITTI, nuScenes and MIAS-LCEC-TF70 benchmarks validate the effectiveness of our method. Remarkably, our approach outperforms prior works on nuScenes using only single-frame point clouds, whereas most existing methods rely on temporally accumulated data.


\clearpage
\setcounter{page}{1}

{
   \newpage
       \twocolumn[
        \centering
        \Large
        \textbf{Single-Frame Point-Pixel Registration via Supervised Cross-Modal Feature Matching}\\
        \vspace{0.5em}Supplementary Material \\
        \vspace{1.0em}
       ]
}

\subsection{Computational Cost and Runtime Analysis}

\subsubsection{Preprocessing}
We report the runtime of the point cloud projection step under different LiDAR configurations, as shown in Table~\ref{tab:preprocessing_runtime}. The number of input points ranges from tens of thousands to several million, corresponding to different LiDAR resolutions and accumulation durations. The results demonstrate that the preprocessing cost scales almost linearly with the number of input points. In practice, projecting the point cloud to the image space mainly depends on point density: about 3.5~ms for 40k points, 16~ms for 200k points, and 38~ms for 500k points on an AMD~Ryzen~5~5600G with Radeon Graphics. This efficiency ensures that our preprocessing remains lightweight enough for real-time deployment.

\begin{table}[htbp]
\caption{Preprocessing runtime under different LiDAR configurations.}
\centering
\footnotesize
\begin{tabular}{lcc}
\toprule
\textbf{Points} & \textbf{Runtime (ms)} & \textbf{Remarks} \\
\midrule
35k   & 3   & 32-line LiDAR \\
100k  & 8   & 64-line LiDAR \\
500k  & 73  & Mid-70 (5s accumulation) \\
2500k & 221 & Mid-70 (20s accumulation) \\
\bottomrule
\end{tabular}
\label{tab:preprocessing_runtime}
\end{table}

\subsubsection{Inference}
Table~\ref{tab:inference_runtime} summarizes the inference runtime and GPU memory usage across different datasets and input resolutions, evaluated on an NVIDIA RTX~4090 GPU. Unless otherwise specified, all experiments resize the longer image side to 840 pixels. The GPU memory requirement varies with both input resolution and projection density. At 840$\times$560 resolution, dense projections run in about 138~ms per pair and consume approximately 4,560~MB of GPU memory, while at 840$\times$472 resolution, sparse projections take around 127~ms and require roughly 2,689~MB. These results show that our approach achieves near real-time performance across different resolutions and projection types, meeting the requirements of online calibration and localization in autonomous systems.

\begin{table}[htbp]
\caption{Inference runtime and memory consumption across datasets and resolutions.}
\centering
\footnotesize
\begin{tabular}{lccc}
\toprule
\textbf{Dataset} & \textbf{Resolution} & \textbf{Runtime (ms)} & \textbf{Memory (MB)} \\
\midrule
nuScenes   & 840$\times$472  & 126.4 & 4,560 \\
\midrule
\multirow{2}{*}{KITTI} 
           & 840$\times$255  & 105.9 & 2,689 \\
           & 1241$\times$376 & 131.5 & 5,490 \\
\midrule
\multirow{2}{*}{MIAS-LCEC} 
           & 840$\times$560  & 137.7 & 5,557 \\
           & 1200$\times$800 & 195.8 & 13,407 \\
\midrule
General    & 840$\times$840  & 166.7 & 9,043 \\
\bottomrule
\end{tabular}
\label{tab:inference_runtime}
\end{table}

\subsection{Feature Matching}
To isolate the effect of modality cues from the sparsity of LiDAR measurements, we use dense point clouds by accumulating 20 seconds of Livox Mid-70 scans. Thanks to the non-repetitive scanning pattern of Livox, this accumulation produces approximately 2500k points, yielding highly dense LiDAR Intensity Projection (LIP) and LiDAR Depth Projection (LDP). Based on these dense projections, we conduct a controlled comparison with MindVision SUA202GC images.

When using intensity (LIP with Gray), the model demonstrates robust and accurate cross-modal correspondences, as indicated by the clear green matching lines. Intensity provides fine-grained reflectance details that enable reliable texture mapping across modalities within a shared feature space. In contrast, depth-based projections (LDP with Gray) lead to less stable results, with more mismatches highlighted in red. This is because depth emphasizes large textureless regions, making it difficult to establish precise correspondences during training.

This dense matching comparison highlights the superiority of intensity over depth as the modality bridging cue in our detector-free registration pipeline, enabling more accurate cross-modal feature alignment.

\begin{figure}[H]
    \centering
    \includegraphics[width=0.95\linewidth]{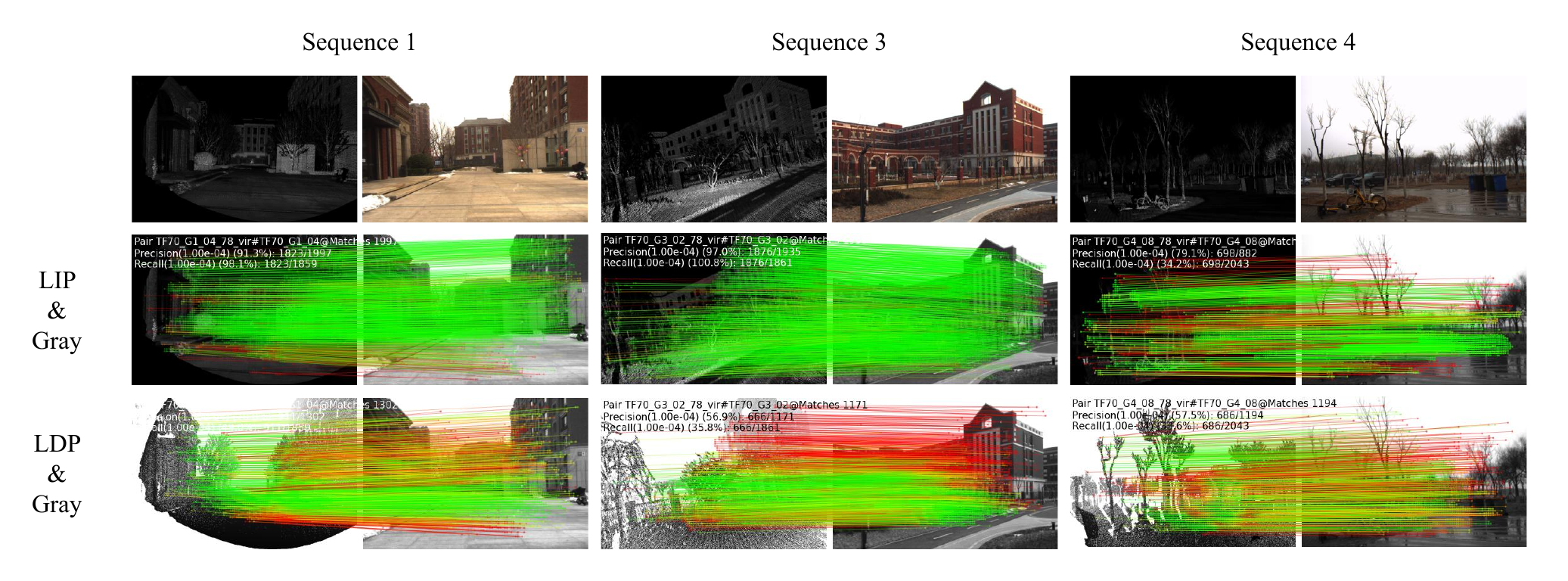}
    \caption{Visualization of cross-modal feature matching between LIP, LDP, and images.}
    \label{fig:vis_s1}
\end{figure}

\subsection{Pose Recovery with EPnP and RANSAC}
To recover the 6-DoF transformation between the LiDAR and the camera, we adopt a robust Perspective-n-Point (PnP) framework. Specifically, we employ the Efficient PnP (EPnP) algorithm~\cite{lepetit2009ep} to compute the camera pose from a set of 2D–3D correspondences. Given matched feature points between the LiDAR projections and the image, each 2D image coordinate is associated with its corresponding 3D LiDAR point. EPnP efficiently estimates the camera pose by solving a linear system based on virtual control points, which significantly reduces computational complexity compared to traditional PnP solvers.

Since the set of correspondences may contain mismatches, we further incorporate RANSAC~\cite{fischler1981random} to ensure robustness. In each iteration, RANSAC randomly selects a minimal subset of correspondences, applies EPnP to generate a pose hypothesis, and evaluates it by measuring the reprojection error over all candidate points. The hypothesis with the largest inlier count is retained as the final solution. This procedure effectively suppresses the influence of outliers and improves the reliability of the estimated transformation. 

It is worth noting that EPnP requires at least four 2D–3D correspondences. Therefore, sample pairs that do not satisfy this requirement are discarded in the computation of average pose error, but are still counted when reporting accuracy (as failed cases).

Through this EPnP+RANSAC pipeline, we obtain a stable and accurate estimation of the extrinsic parameters \(\boldsymbol{T}\), which aligns the LiDAR and camera coordinate frames. The robust estimation not only compensates for potential mismatches in feature correspondences but also guarantees precise cross-modal registration in subsequent experiments.

\subsection{Qualitative Comparison}

\begin{figure}[H]
    \centering
    \includegraphics[width=1.00\linewidth]{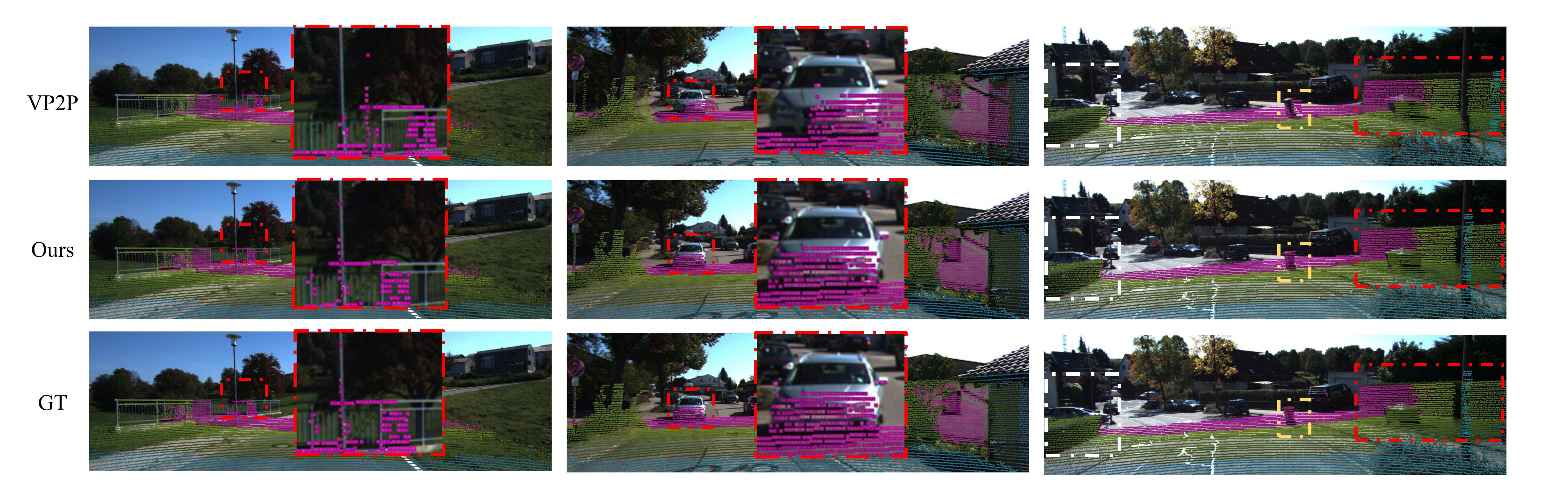}
    \vspace{-1.8em}
    \caption{Visual comparison of Image-to-Point Cloud registration results under the KITTI Odometry dataset.}
    \label{fig:vis}
\end{figure}

We further provide a visual comparison in Fig. \ref{fig:vis}. The figure presents qualitative results of our method in comparison with VP2P and the ground truth. For each example, key regions are highlighted with bounding boxes, and the first two columns show zoomed-in views. The last column displays the complete point cloud projections, where it can be clearly observed that our method achieves superior performance in semantically rich regions of the scene, with pose estimations that are closely aligned with the ground truth. The improved performance can be attributed to the characteristics of LiDAR data, which is typically dense in the near range and sparse in the far range. When the nearby region contains sufficiently informative intensity measurements, our method is able to utilize this information effectively.

\end{document}